\documentclass{edm_article}
\usepackage{amsmath,graphicx,hyperref}
\usepackage{xcolor}
\usepackage{enumitem}
\usepackage{url}
\usepackage{multicol}  % For two-column layout
\usepackage{tcolorbox} % For creating text boxes
\usepackage{booktabs}

\begin{document}

\title{Improving Speech Recognition of Named Entities in Classroom Speech with LLM Revision and Phonetic-Semantic Context}

% Submissions for EDM are double-blind: please do not include any author names or affiliations in the submission. 
% Anonymous authors:
\numberofauthors{3}
\author{
	\alignauthor
Viet Anh Trinh\\
       \affaddr{Worcester Polytechnic Institute}\\
       \email{anhtv1@gmail.com}
\alignauthor
Xinlu He\\
       \affaddr{Worcester Polytechnic Institute}\\
       \email{xhe4@wpi.edu}
\alignauthor
Jacob Whitehill\\
       \affaddr{Worcester Polytechnic Institute}\\
       \email{jrwhitehill@wpi.edu}
}

\maketitle

\begin{abstract}
Classroom speech and lectures often contain named entities (NEs) such as names of people and special terminology. While 
automatic speech recognition (ASR) systems have achieved remarkable performance on general speech, the word error rate (WER) of state-of-the-art ASR remains high for named entities. Since NE are often the most critical keywords, misrecognizing them can affect all downstream applications, especially when the ASR functions as the front end of a complex system. In this paper, we introduce a large language model (LLM) revision pipeline to revise incorrect  NEs in ASR predictions by leveraging not only the LLM's world knowledge and reasoning ability but also  the available phonetic and semantic context. We also introduce the NER-MIT-OpenCourseWare dataset, containing 45 hours of data from MIT courses for development and testing. On this dataset,  our proposed technique achieves up to 30\% relative WER reduction  for NEs. 
\end{abstract}

\section{Introduction}
In the past several years, automatic speech recognition (ASR) performance has improved significantly by scaling data and model size. Whisper \cite{radford2023robust},  which was trained on 765K hours of speech, has achieved remarkable performance, achieving a WER of 2.7\% on the standard Librispeech clean corpus and an average of 12.8\% across popular speech datasets \cite{radford2023robust}. However, Whisper (large-v3) still falls short in recognizing named entities (NEs). For instance, it has a WER of  28.9\% for persons and 19.6\% for organizations in the contextual ASR benchmarking dataset ConEC \cite{huang2024conec}. Since classroom and academic speech often contains NEs, the inability to recognize them correctly can cause downstream problems in various educational applications, such as topic modeling of curricula \cite{rouly2015we,takizawa2023using,amjad2022advanced}, interactions with real-time educational agents \cite{d2024learning,lee2025collaborative}, and more.

Humans, on the other hand, have the ability to reason and relate different NEs based on knowledge and local context. For example, students can use lecture slides to map the NEs in the speech from a lecture video to the content in the slides, using both semantic  as well as phonetic information. Since  NEs relate to events, persons' names, products, etc.~which often depend on local context and change over time, the solution of retraining a model with new paired speech-text data is expensive, and hard to scale.

\begin{figure*}[t]
\begin{center}
    \includegraphics[width=.8\textwidth]
    {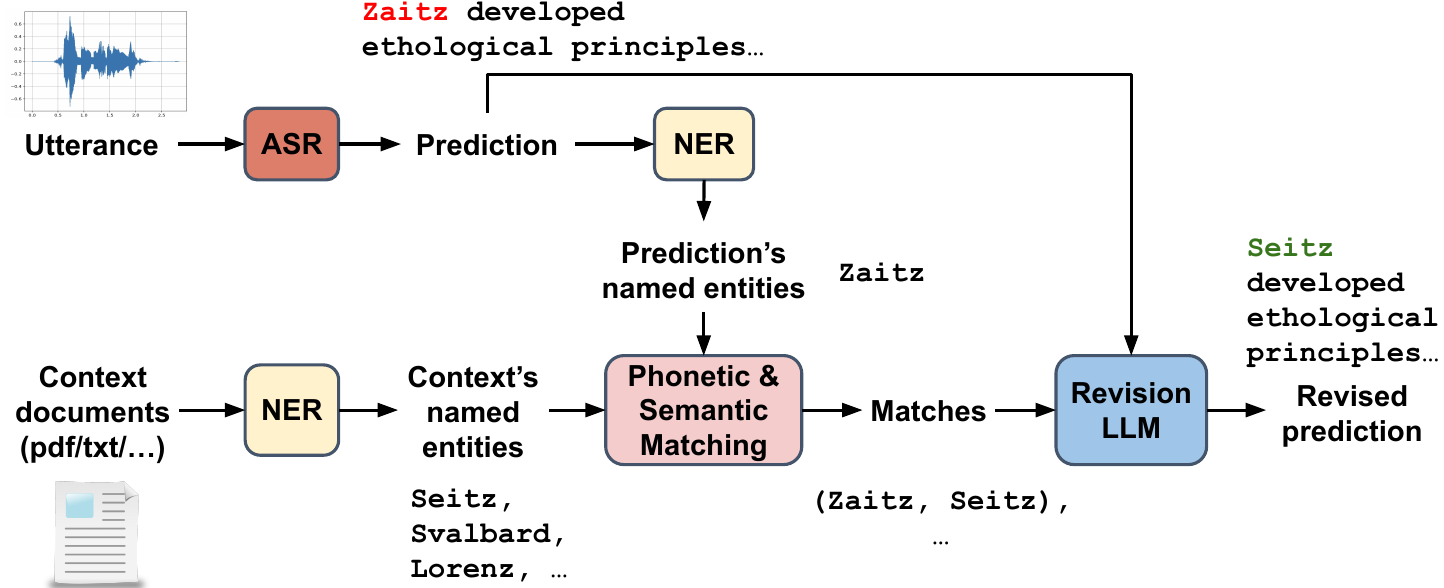}
    \end{center}
    \caption{Proposed method of combining phonetic and contextual information to revise named entities in ASR predictions.}
    \label{fig:aiagent}
\end{figure*}

{\bf Contributions}: (1) We propose a novel approach (see Fig.~\ref{fig:aiagent}) that utilizes an LLM to revise incorrect NEs in ASR transcripts. 
Specifically, the LLM receives predictions from the ASR and filtered information from the local context, which contains similar-sounding NEs and sentences that include these words, and uses this information to revise the ASR predictions.
In contrast to prior works on LLM revision of ASR, our method harnesses combines LLM-based reasoning and world knowledge with explicit phonetic and semantic context, without needing to be trained on a specific NE list. The method is interpretable because the LLM explains why it makes certain revisions, unlike other attention-based approaches that only output the revision without any explanation. (2)
We demonstrate that our proposed method works effectively with different LLMs, ranging from closed-source to open-source models and across various sizes.
%Additionally, our work can be extended in the future by incorporating visual information or addressing the error rate for rare words.
(3) Finally, we introduce the MIT-OpenCourseWare NER contextual dataset, which includes both speech-text pairs and local contexts\footnote{\url{https://huggingface.co/datasets/lucille0he/ocw}}.
%
%In this dataset, our proposed method achieves the relative WER reduction on named entities word error rate (NE-WER) of 30 \%. 
%In addition, we also release the MIT-OpenCourseWare dataset for ASR NER performance purpose. The dataset contains of x hours from the MIT-OpenCourseWare \cite

%JAKE: talk about LLM agents and related applications...

\section{Related Work}
{\bf Fine-tuning Speech Foundation Models for Classroom Speech}:
The past few years have seen growing interest in fine-tuning speech foundation models (e.g., wav2vec2 \cite{baevski2020wav2vec} and WavLM \cite{chen2022wavlm}) for classroom-specific recording conditions \cite{southwell2024automatic,attia2024continued}. While this approach can increase the average transcription accuracy, it does not focus on named entities in particular. Even though NEs may occur more frequently in classroom speech than in everyday speech, they still make up only a small fraction of the overall audio recording; hence, the ability of fine-tuning to improve NE recognition performance is limited. In contrast, our proposed method is ``zero-shot'' in the sense that it uses an pre-trained LLM's ability to to fix errors, given enough phonetic and semantic context, without needing a large dataset for parameter tuning.

{\bf Biasing}: A common way to improve NE recognition is by biasing the ASR toward a list of predefined words.  This can be implemented with shallow fusion \cite{kannan2018analysis, he2019streaming, zhao2019shallow, mohri2002weighted} whereby,  during inference, the ASR prediction is influenced by a contextual language model with a certain weight. In practice, such systems are sensitive to the weight parameter. More powerful are biasing approaches
that combine the context and acoustic information more tightly, e.g., with a neural network that encodes contextual information and is trained jointly with the ASR system \cite{pundak2018deep, chang2021context, sathyendra2022contextual, gong2024advanced}. One of the difficulties in biasing approaches is that performance often decreases as the number of approved NEs grows. Our proposed solution addresses this issue by selecting only the NEs in the context that sound similar to those in the ASR prediction.

{\bf LLM-based ASR revision}: During the past two years, there has been significant interest in developing LLM-based methods to revise ASR hypotheses \cite{peyghan2025survey}.
For instance, \cite{pu2023multi} use language model rescoring followed by an LLM in zero-shot fashion to  revise the ASR hypothesis based on a ranked list of most
likely candidates. \cite{rao2025} also use an LLM to revise the ASR hypotheses based on the most likely candidates but additionally
impose constraints so that the revised prediction is similar to the original one in terms of Levenshtein distance. Neither of these methods harness
semantic context or phonetic similarity. Most similar to ours is work by
\cite{wang-etal-2024-dancer}, which combines phonetic and contextual information to revise NEs in ASR transcriptions. 
However, in contrast to our method, which is training-free and generalizable to different vocabularies, their
approach requires training a masked language model on a fixed set of NEs to detect NE mistranscriptions.

%Neural networks are often used to encode contextual information \cite{pundak2018deep, chang2021context, sathyendra2022contextual, gong2024advanced}. They are typically trained jointly with the ASR system. Early work by \cite{pundak2018deep} introduced a bias encoder to the encoder-decoder ASR model \cite{chan2015listen}. In \cite{chang2021context}, a neural network is used to encode the bias words, and a biasing layer fuses the contextual network embedding with the audio encoder or label encoder of a transducer-based ASR using attention. \cite{le2021contextualized} combines trie-based biasing, shallow fusion, and contextual neural networks.  \cite{han2021cif} proposes collaborative decoding with a continuous integrate-and-fire model for contextual ASR. Recently, there has been some work related to contextual speech recognition that utilizes LLMs \cite{gong2024contextual}. In \cite{gong2024contextual}, the authors propose using a biasing prompt and a biasing fusion network.  %For the biasing prompt, the authors suggest adding special tokens to identify the start and end of bias words.  This work also highlights that increasing the length of the bias list significantly reduces performance.  Our proposed solution addresses this issue by selecting only the named entities in the context that sound similar to those in the ASR prediction. This approach reduces the length of the bias list while still retaining the most relevant bias words.

%-----------------
%-----------------

% Finish checking grammar till here
\section{Proposed Method: Phonetic \& Contextual LLM Revision}
We propose a novel method of using an LLM's reasoning abilities to revise ASR predictions by harnessing \emph{contextual} as well as  \emph{phonetic} information on named entities in both the speech and the available context. The method consists of several steps. First, (1)
speech is transcribed using an ASR model (e.g., Whisper).  Next, (2) NEs are extracted from both the ASR predictions and the context documents using an off-the-shelf NE recognition tool. The context can be any text document related to the speech.  For instance, when transcribing lectures, the context might include slides or reading lists. Due to the large number of words in the context, (3)  the NEs are selected from the context only if they share the same entity type (e.g., person, organization) and sound similar to the NEs in the ASR predictions. 
Finally (4),  an LLM analyzes the ASR predictions and their NEs, as well as the NEs in the context and the sentences containing them. It uses this information and a detailed prompt of instructions (see Fig.~\ref{fig:prompt}) to produce a revised prediction.

%--------------------------------
%\SetAlgoNoLine
%\DontPrintSemicolon
%\LinesNumbered
%\begin{algorithm}
%\SetKwFunction{revision}{SpeechRevision}
%\SetKwProg{Fn}{Function}{:}{}
%\Fn{\revision{ASR prediction $h$, context $c$}}{
%    \tcp{Extract named entities from the context and prediction}
%    NamedEntitiesContext $\gets$ \texttt{NER}($c$)\;
%   NamedEntitiesPrediction $\gets$ \texttt{NER}($h$)\;
    
%    \tcp{Find entities with similar sound and same type}
%    \If{\texttt{SoundSimilar}(NamedEntitiesContext, NamedEntitiesPrediction) \textbf{and} \texttt{SameType}(NamedEntitiesContext, NamedEntitiesPrediction)}{
%        SimilarSoundEntities $\gets$ NamedEntitiesContext\;
%    }
    
%    \tcp{Prepare context for LLM and revise the prediction}
%    ContextLLM $\gets$ SimilarSoundEntities\;
%    RevisedPrediction $\gets$ \texttt{LLM}(ContextLLM, $h$)\;
    
%    \KwRet RevisedPrediction\;
%}
%\caption{Speech Revision Algorithm}
%\label{alg:alg1}
%\end{algorithm}

%------------------------------
\section{Dataset}
\label{sec:dataset}
As existing datasets for contextual ASR are very limited and often tailored to specific industry settings (e.g., earnings calls), we collected our own dataset on topics of wider scientific and educational interest. In particular, we constructed a new dataset, called NER-MIT-OpenCourseWare, based on content from MIT OpenCourseWare. We chose this educational repository as a basis for our dataset because of the availability of detailed context  documents along with high-quality transcriptions of the speech content.

\begin{figure*}[t]
\centering
\setlength{\fboxsep}{0pt} % remove padding inside box
\setlength{\fboxrule}{0.5pt} % thickness of the frame
\fbox{\includegraphics[angle=90, width=.9\columnwidth]{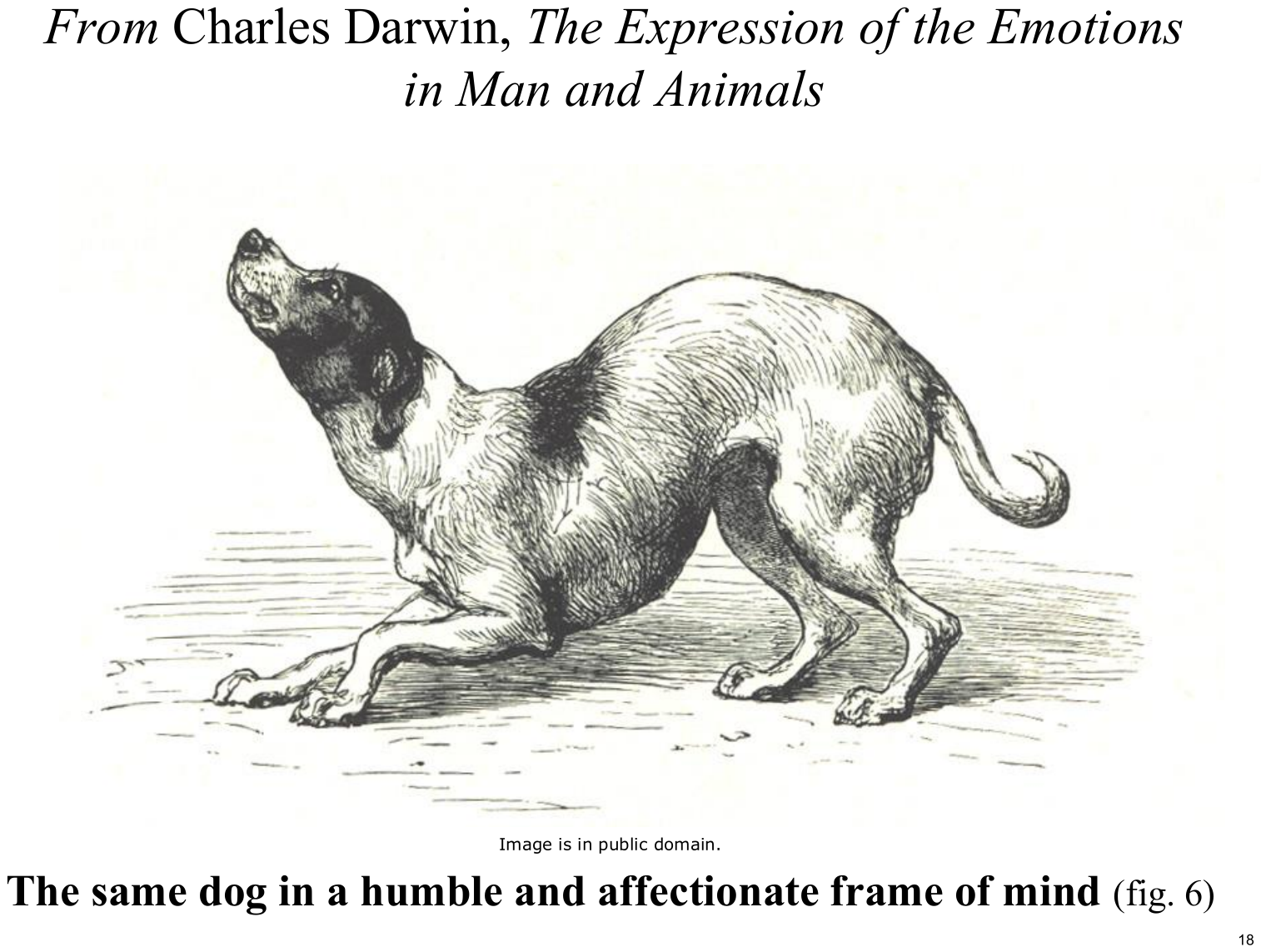}}
\fbox{\includegraphics[angle=90, width=.9\columnwidth]{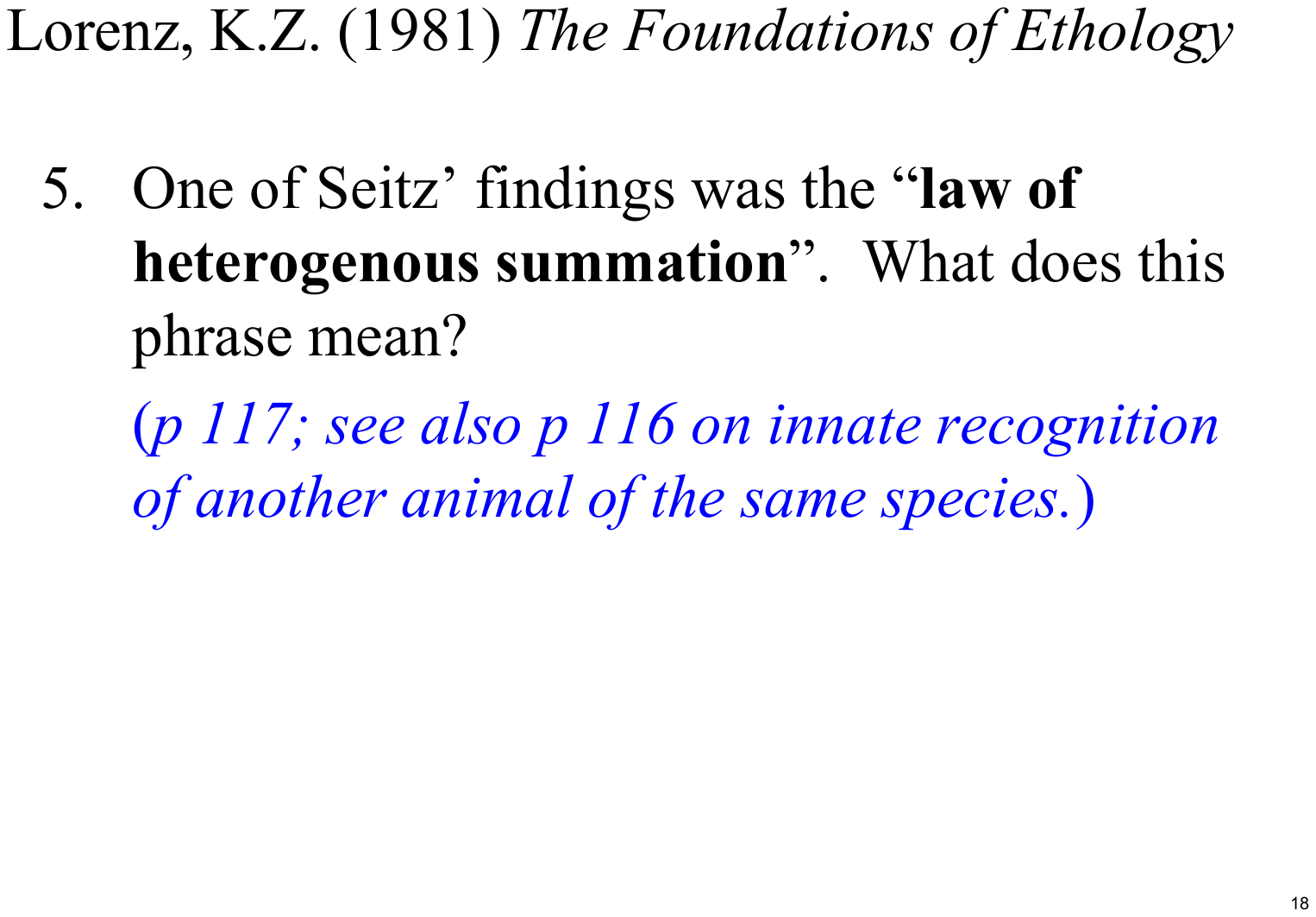}} \\ %&
\fbox{\includegraphics[angle=90, width=.9\columnwidth]{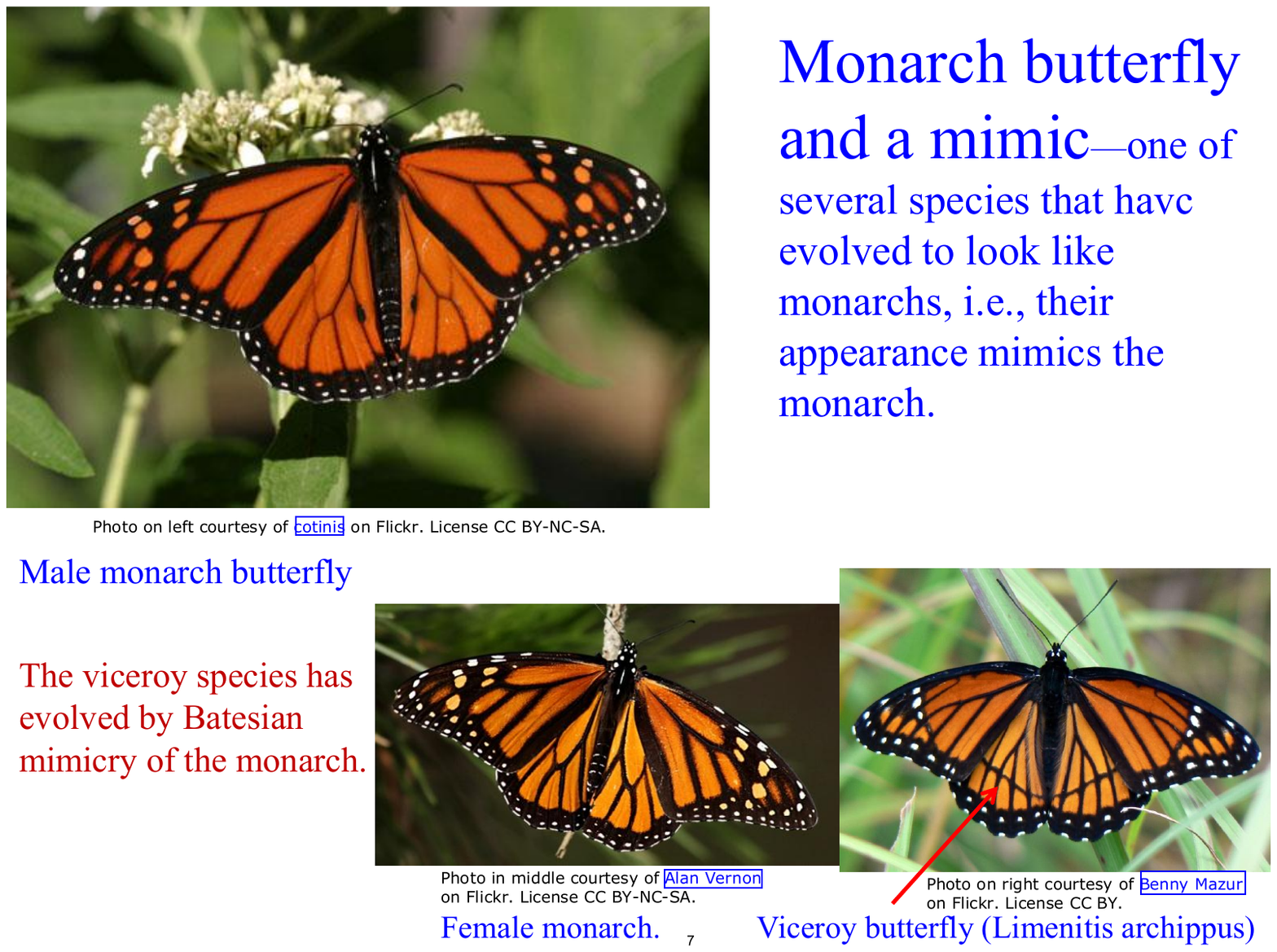}}
\fbox{\includegraphics[angle=90, width=.9\columnwidth]{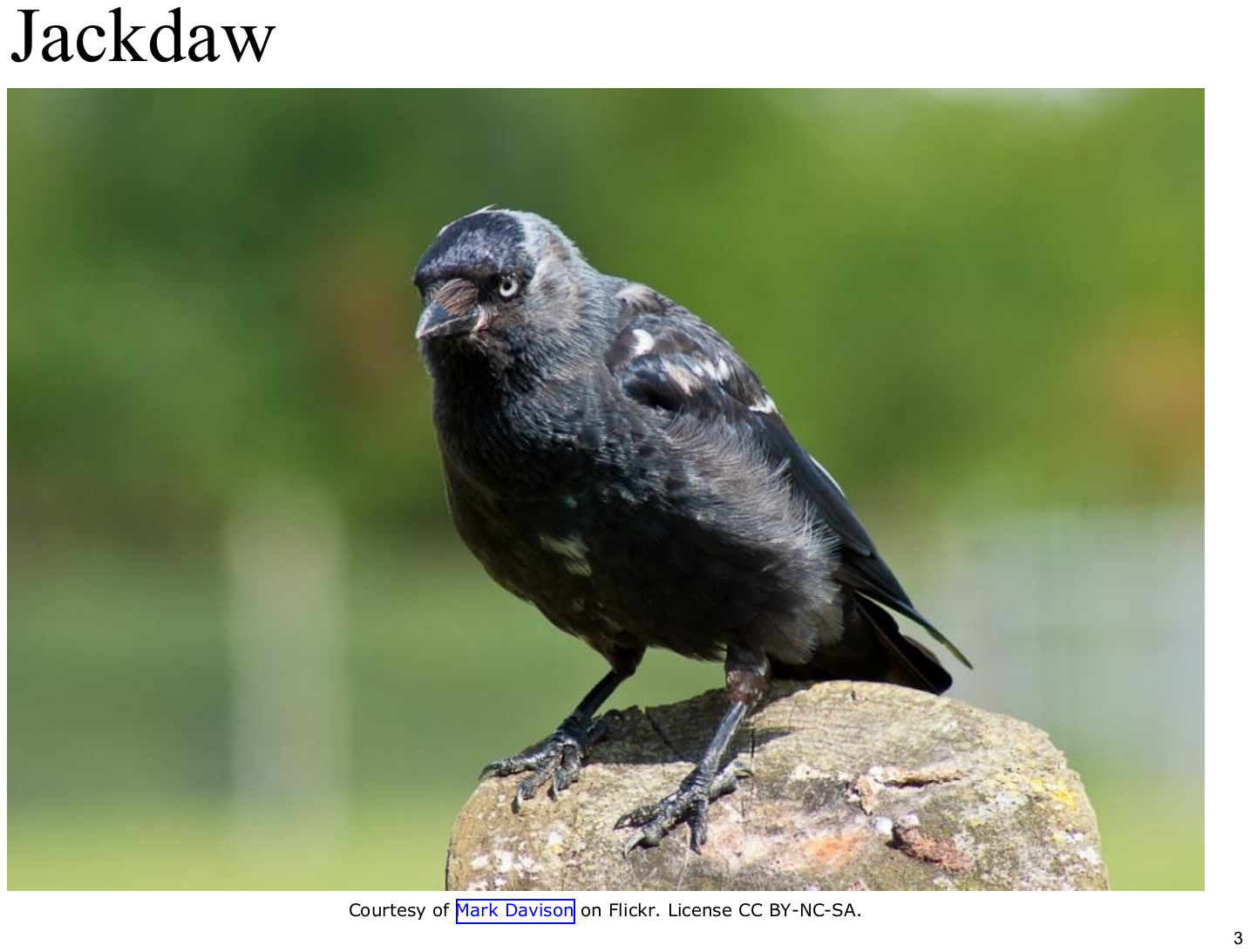}} \\
\caption{Examples of NER-MIT-OpenCourseWare slides.}
\label{fig:slides}
\end{figure*}

For the  test set, we used the lectures from an undergraduate course on animal behavior, offered in Fall 2013 from MIT OpenCourseWare, containing  25 lecture audio recordings  totaling 17 hours of audio.
 The test set labels contain approximately 121K words, while the context consists of  46K words.  The ground-truth transcriptions contain  1.2K NE words, and about 66\% of these words appear in the slides. Examples of lecture slides are shown in Figure \ref{fig:slides}. 

For the development set, we chose an undergraduate course about brain structure, offered in Spring 2014. The course has 35 lectures, but we only use the first 3K utterances to enable faster execution. The labels contain approximately 32K words, while the context includes 14K words from six lecture slides. There are 0.2K NE words in the development set, and around 60\% of these words appear in the slides.

We used BeautifulSoup\footnote{\url{https://www.crummy.com/software/BeautifulSoup}} to download the audio, video, and lecture content;
pdfplumber\footnote{\url{https://github.com/jsvine/pdfplumber}} to convert the ground-truth transcripts and the slides into text files; and spacy \begin{tt}en\_core\_web\_lg\end{tt} to remove unwanted characters and split the transcripts into sentences. The slides contain images, math equations, and other elements, making them noisy. We used pydub\footnote{\url{https://github.com/jiaaro/pydub}} to extract audio (flac) from video (mp4) and convert the audio to mono with a sampling rate of 16KHz. Since each audio file has a duration of 30-60 min, we used fairseq \cite{ott2019fairseq} to segment each audio into utterances corresponding to the sentences identified by the spacy model.

%-----------------------------
%-----------------------------------------
\section{Experiments}
\label{sec:experiments}
Using NER-MIT-OpenCourseWare (Sec.~\ref{sec:dataset}), we conducted experiments to assess the benefits of the proposed method (Fig.~\ref{fig:aiagent}) to improve ASR transcription accuracy of NEs, while hopefully keeping the transcription accuracy of non-named entities (NNEs) unchanged. 
As our method focuses on filtering relevant information in the context using contextual and phonetic cues before feeding it to the LLM, we compare against three strong control conditions:
(1) no LLM, just phonetic matching of the predicted NE with a randomly chosen similar sounding word of the same entity type from the context; 
(2) giving the revision LLM the entire context document corresponding to a given utterance (not just the filtered relevant entities); and (3) first using an LLM to summarize the context, and then providing the summary to the revision LLM.

\subsection{Implementation Details}
For the first-stage ASR,  we use  Whisper large-v3. We choose Flair \cite{akbik2019flair}, specifically the flair/ner-english-ontonotes-large model, for named entity recognition (NER) due to its speed and high accuracy on standard NER benchmarks.
To detect whether two words sound similar to each other, we use Double Metaphone \cite{philips2000doublemetaphone}, which is 
a popular rule-based algorithm that converts words into phonetic codes that is useful for detecting similar-sounding words and name searches \cite{blair2024jrc}. Specifically, Double Metaphone outputs one primary and one alternative phonetic code (for names that may be pronounced in different ways). In our proposed method, we defined that two words sound ``similar'' to each other if either their primary phonetic codes or equivalent or their alternative phonetic codes are equivalent. %, trias2021named}.
We note that the use of Double Metaphone could conceivably be ``swapped out'' for another phonetic matching system (e.g., one based on multimodal LLMs with shared semantic and acoustic embeddings) to work for other languages. We leave this to future work.
%It can be applied not only to English but also to low-resource languages \cite{luitel2024contextual} %, kunchukuttan2021large}. 

Similar to prior contextual ASR research \cite{huang2024conec}, we focus on eight specific NE types: PERSON, ORG, GPE, LOC, PRODUCT, EVENT, NORP, and FAC. We manually verify the NEs in 500 sentences from the test set and measure set-based evaluation recall, ignoring multiple occurrences of the same NE in a sentence, as the LLM replaces all instances of the same NE in the sentence. The precision of Flair is 96\%, and the recall is 86\%. Since manually labeling NEs in the dataset is expensive, and Flair's recall rate was reasonably high, we use Flair's detected NEs as a basis for computing WER. %ground-truth to evaluate the WER on Flair-detected NEs.

%\begin{table*}[h]
\begin{table}
\centering
\caption{Comparison of named entity (NE) word error rates (WER) (\%) and non-named entity (NNE) WER (\%) across methods using Whisper-large-v3 as the ASR.}
\begin{tabular}{llcclcclcc}
\toprule
\textbf{Revision Method}            & \textbf{LLM} & \textbf{NE} & \textbf{NNE} \\ 
\midrule
None                   & --                   & 32.3     & 7.0 \\ \hline
Phonetic matching      & -- & 25.3 & 7.0 \\ \hline
Full context           & GPT-4o-mini          & 43.4    & 7.1 \\
& GPT-4o               & 30.6   & 7.0  \\
& Llama-3.1-70B        & 32.6   & 7.2      \\  \hline
Context summary        & GPT-4o-mini          & 38.6  & 7.1 \\
& GPT-4o               & 54.2  & 7.1 \\
& Llama-3.1-70B        & 47.4  & 8.2                  \\  \hline
Proposed method        & GPT-4o-mini          & 22.7  & 7.0      \\
& GPT-4o               & 22.9   & 7.0  \\
& Llama-3.1-70B  & 24.8  & 7.2   \\
\bottomrule
\end{tabular}
\label{tab: multiple_llms}
\end{table}

For the revision LLM, we compare Llama3-70b-8192  (via Groq API), GPT-4o-mini API, GPT-4o API , and Meta-Llama-3.1-70B (using bitsandbytes quantization), running locally on a single A100 GPU (80GB) with vLLM \cite{kwon2023efficient}. %Langchain\footnote{\url{https://github.com/langchain-ai/langchain}} is chosen as the framework for the experiment, as it provides flexibility to switch between different LLM providers. 
We draw inspiration from \cite{brown2020language} %and the OpenAI Prompt Engineering Guide\footnote{\url{https://platform.openai.com/docs/guides/prompt-engineering}} 
to design the prompt. Our prompt (see Fig.~\ref{fig:prompt}) is designed to ensure that the LLM revises NEs without altering other words and applies reasoning.  %Because the LLM outputs additional text beyond the revised prediction, we instruct it to use a special format for the revised prediction, starting with $<<@$ and ending with $@>>$. This approach allows us to easily separate the revised prediction from the additional text and has proven robust across all LLMs.
The asr\_confidence\_threshold used in the prompt was optimized on the development set.

%-----------------------
\begin{figure}[t]
\centering
\fbox{%
  \parbox{.95\linewidth}{
\label{box:prompt}
\footnotesize
Speech recognition may incorrectly capture named entities due to similar-sounding words or uncommon entities in training data.
                The named entities and their probabilities are provided below as "Named entities probability" in the format: entity : probability.
                Review the prediction's named entities. For each entity whose probability is below {asr\_confidence\_threshold}, or missing, please complete the following steps:
                \begin{itemize}[leftmargin=*]
                \item Examine the context to determine what the similar-sounding entity refers to, and analyze the sentence to understand what the ASR-predicted entity refers to. 
                Check if (1) both refer to the same thing, (2) the difference is likely due to an ASR transcription error, (3) the replacement improves accuracy, preserves the original sentence's meaning, and fits naturally within the sentence.
                If yes to those 3 criteria, then replace the ASR-predicted entity with the similar-sounding entity.
                \item Make sure to treat each named entity as a cohesive unit when evaluating or replacing. Do not modify individual words separately %with
                in a multi-word entity.
                \item If no suitable similar-sounding entity exists, leave the entity unchanged.
                \item Do not delete any words in the entity that are absent from the context.
                \item Keep all other words, punctuation, and formatting unchanged.
                \end{itemize}
                The context includes similar-sounding entities and sentences containing these named entities.
                The revised prediction should start with $<<@$ and end with $@>>$
                Context: \{context\}
                
                Speech recognition prediction: \{preds\}
                
                Named entities probability: \{ent\_prob\_dict
                \}
  }
}
\normalsize
\caption{ASR revision prompt given to the LLM. In an ablation analysis, we omit the second, third, fourth, and fifth bullets points.}
\label{fig:prompt}
\end{figure}

For the summarization task (in control condition 2), we use GPT-4o-mini
with the prompt: ``Summarize the following context concisely while ensuring all named entities are preserved. Include a sentence for each named entity, using only alphabets and digits. Context:\{context\}''.

%-----------------
\begin{table}
    \centering
    \caption{Comparison of named entity (NE)  and non-named entity (NNE) WER (\%) across speech recognition systems utilizing GPT-4o-mini in our proposed method.}
    \label{tab:wer-canary}
    %\begin{tabular}{@{}llc@{}c@{}}
    \begin{tabular}{@{}llcccc@{}}
        \toprule 
\textbf{ASR} & \textbf{Params} & \multicolumn{2}{c}{\textbf{Before revision}} & \multicolumn{2}{c}{\textbf{After revision}} \\ 
             &                 & \textbf{NE} & \textbf{NNE}                & \textbf{NE} & \textbf{NNE} \\ 
        \midrule
        Whisper-sm                  & 244 M                                                & 39.3 & 9.4 & 29.5 & 9.4                      \\
        Whisper-me          & 769 M & 36.0 & 8.0 & 26.8 & 8.0                        \\
        Whisper-la       & 1.54 B & 32.3 & 7.0 & 22.7  & 7.0                         \\
        Canary-1B          & 1B  & 36.6 &  8.3 & 26.8   & 8.3                        \\ \bottomrule
    \end{tabular}
\end{table}

\subsection{Results}
As shown in Table \ref{tab: multiple_llms},
the proposed method demonstrates improved accuracy compared to the original Whisper predictions, regardless of which LLM was used. 
In particular, the proposed approach reduces the WER of NE to 22.7\% (corresponding to a 30\% relative WER reduction) while maintaining the NNE WER at 7\%.
Phonetic matching of NEs to the context (control condition 1) is also useful and decreases the WER to 25.3\%.
In contrast, when using an LLM with the full context (control condition 2) without any filtering,
the WER actually increases from 32.3\% to 43.4\%.  This could be due to the long prompt input, which causes the LLM to hallucinate:
a follow-up analysis of the animal behavior course test set revealed that
the full-context condition using GPT-4o resulted in statistically significantly longer revisions, on average, compared to the proposed method
($t(9426)=4.11, p<0.0001$).
%In [101]: scipy.stats.ttest_1samp(df_merged.word_diff.to_numpy(), 0)
%Out[101]: TtestResult(statistic=np.float64(4.113424918927835), pvalue=np.float64(3.9314504226414375e-05), df=np.int64(9426))
Control condition 3, which uses a summary of the full context, is also ineffective, as the WER increases from 32.3\% to 38.6\%.
Together these results indicate that, despite being trained on large amounts of text data, the LLM's world knowledge and linguistic reasoning are insufficient to improve the ASR accuracy of NEs. Instead, both a filtered context as well as explicit phonetic information are necessary to enable the LLM to improve NE transcription.

As a follow-up analysis, we also computed the WER of the proposed method (using GPT-4o-mini and Whisper) on named entities (NEs) not included in the context documents: it is 23.6\%. This is only slightly higher than the overall WER on NEs (22.7\%), and it is still substantially lower than the WER on NEs with just Whisper alone (32.3\%). This suggests that the proposed method is unlikely to do harm, even if its benefits are more pronounced on NEs that are contained in the context.

%Fig.~\ref{fig:sample_revised_prediction} shows a qualitative example of how our proposed method effectively fixes mistakes by various ASR systems. 

Our proposed method also demonstrates its effectiveness across different LLM models, such as the large, state-of-the-art GPT-4o and the robust open-source Llama-3.1-70B, as shown in Table \ref{tab: multiple_llms}. For a powerful model like GPT-4o, utilizing the full context results in a slight WER reduction, from 32.3\% to 30.6\%, as indicated in Table \ref{tab: multiple_llms}. However, with our proposed method, the WER is significantly reduced to 22.9\%.  

To evaluate the effectiveness of the proposed approach over different ASR systems, we conducted experiments with Whisper-small, Whisper-medium, Whisper-large-v3, and Nemo Canary-1B \cite{kuchaiev2019nemo}, another state-of-the-art ASR model. As shown in Table \ref{tab:wer-canary}, our  method generalizes well across different speech models: With Canary-1B, the NE WER is reduced from 36.6\% to 26.8\%.  Similarly, the NE WER is reduced from 39.3\% to 29.5\% with Whisper small and from 36.0\% to 26.8\% with Whisper medium. 

Finally, we also experiment with a simpler version of the prompt that omits the extra guidelines contained in the second, third, fourth, and fifth bullet points  of Figure \ref{fig:prompt}. This shortened prompt's results are reported in Table \ref{tab:ablation-1}. As shown, both the full prompt and the short prompt perform well with GPT-4o-mini, while the open-source Llama-3.1-70B achieves better performance with the full prompt at 24.8\% compared to the short prompt at 32.3\%. This indicates that the paid API GPT-4o-mini may be more robust and have better reasoning capabilities than the open-source Llama-3.1-70B in this case. The open-source model appears to require more detailed instructions to achieve good results, whereas GPT-4o-mini can perform well without them.

\subsection{Examples}
The ground-truth transcript (converted to lower case) of one utterance is, ``now this is a difficult concept but let me try to go through it anyway because \color{blue}seitz \color{black} was very important in working out some of the principles of ethological investigation and the principles of ethology that   \color{blue}lorenz \color{black} was talking about''. Canary 1B transcribed it as, ``now this is a difficult concept \ldots \color{red} zeitz \color{black} \ldots \color{red} lawrence \color{black} \ldots'' (where the ellipses indicate correctly transcribed words that we omit for brevity). Whisper transcribed it as, ``now this is a difficult concept\ldots \color{red} zaitz \color{black} \ldots \color{red} lawrence \color{black} \ldots''. After LLM-based revision with the proposed method, both Canary and Whisper correctly transcribed it as, ``now this is a difficult concept \ldots \color{blue} seitz \color{black} \ldots \color{blue}lorenz \color{black} \ldots''.

In another utterance, the ground-truth transcript is, ``one of those students became very well known margaret \color{blue}mead\color{black}''. With phonetic matching (control condition 1), the NE was replaced with a randomly chosen similar-sounding NE, resulting in ``\ldots margaret \color{red}mit\color{black}''. With our proposed method (using GPT-4o-mini), however, the  ASR transcript is correctly transcribed as, ``\ldots margaret \color{blue}mead\color{black}''.

%----------------------
%Additionally, we experiment with optimizing our prompt. Since our approach is designed to generalize across different LLMs and model sizes, we write a prompt that includes as much detail as possible, rather than optimizing it for a specific LLM or model size. 
\subsection{Run-time Cost}
Extracting NEs takes about 0.013sec per utterance (averaged over 9427 test utterances). Revising each ASR prediction using GPT-4o-mini takes about 1.2sec (averaged over 750 test utterances with NEs).
%The revision time for Qwen 2.5 7B is 0.35 second [MAYBE USE LLAMA 8B?]

%\subsection{Phonetic Cues Only}
%Finally, we also conducted an experiment that does not require an LLM, employing a simple strategy of randomly replacing the named entity in the ASR prediction with one of its similar-sounding named entities from the context. On NEs, the WER using random replacement is 25.3\%, compared to 22.7\% using the proposed method with GPT-4o-mini. This is likely because the simpler baseline replaces NEs without understanding the context, which becomes particularly problematic with noisier contextual data. As an example, for an utterance whose ground-truth transcript is ``one of those students became very well known margaret \color{blue}mead\color{black}'', replacing the NE with a randomly chosen similar-sounding entity results in ``\ldots margaret \color{red}mit\color{black}''. With our proposed method (using GPT-4o-mini), however, the  ASR transcript is correctly transcribed as ``\ldots \color{blue}mead\color{black}''.

%\begin{table}[htbp]
\begin{table}
    \centering
    \caption{Named entity WER (\%) and non-named entity WER are evaluated using the full prompt or the short prompt.}
    \label{tab:ablation-1}
    \begin{tabular}{@{}llcc}
        \toprule
        \textbf{Method}            & \textbf{Revision LLM}                                  & \textbf{NE (\%)} 
        & \textbf{NNE} \\ \midrule
        Full prompt                   & GPT-4o-mini                                                & 22.7 & 7.0                        \\
        Full prompt                   & Llama-3.1-70B                                              & 24.8   & 7.2                     \\
        Short prompt         & GPT-4o-mini  & 22.7  & 7.0                        \\
        Short prompt         &Llama-3.1-70B  & 32.3      & 7.0                 \\
        \bottomrule
    \end{tabular}
\end{table}
%----------------------

%-----------------
\section{Conclusions}
We introduced a novel method to revise predictions from ASR using an LLM. Specifically, we employ a phonetic and contextual matching component to select NEs from the context that are most relevant to those in the ASR prediction.
We compare our proposed method to strong LLM controls that use the full context or a context summary, as well as a phonetic matching baseline, and show that our method achieves superior performance -- up to 30\% relative WER reduction on a new and challenging test set based on MIT OpenCourseWare lecture series. %In future work, we could expand our method with new approaches to detect similar-sounding entities based on acoustic and semantic embeddings.
We focused primarily on NEs since their definitions are standardized, and tools for NER are readily available. However, the method could be extended to revise rare words with the development of rare word detection models. Furthermore, the method could be expanded to integrate visual information, such as image captions or optical character recognition, so as to better harness available context.

\subsection*{Acknowledgement}
This research was supported by the NSF National AI Institute for Student-AI Teaming (iSAT) under grants DRL \#2019805 and DRL \#2454151, as well as  from an NSF CAREER grant \#2046505. The opinions expressed are those of the authors and do not represent views of the NSF.

\bibliographystyle{abbrv}
\bibliography{paper}  % sigproc.bib is the name of the Bibliography in this case

@article{chen2022wavlm,
  title={Wavlm: Large-scale self-supervised pre-training for full stack speech processing},
  author={Chen, Sanyuan and Wang, Chengyi and Chen, Zhengyang and Wu, Yu and Liu, Shujie and Chen, Zhuo and Li, Jinyu and Kanda, Naoyuki and Yoshioka, Takuya and Xiao, Xiong and others},
  journal={IEEE Journal of Selected Topics in Signal Processing},
  volume={16},
  number={6},
  pages={1505--1518},
  year={2022},
  publisher={IEEE}
}

@article{baevski2020wav2vec,
  title={wav2vec 2.0: A framework for self-supervised learning of speech representations},
  author={Baevski, Alexei and Zhou, Yuhao and Mohamed, Abdelrahman and Auli, Michael},
  journal={Advances in neural information processing systems},
  volume={33},
  pages={12449--12460},
  year={2020}
}

@inproceedings{southwell2024automatic,
  title={Automatic speech recognition tuned for child speech in the classroom},
  author={Southwell, Rosy and Ward, Wayne and Trinh, Viet Anh and Clevenger, Charis and Clevenger, Clay and Watts, Emily and Reitman, Jason and D’Mello, Sidney and Whitehill, Jacob},
  booktitle={Icassp 2024-2024 ieee international conference on acoustics, speech and signal processing (icassp)},
  pages={12291--12295},
  year={2024},
  organization={IEEE}
}

@article{attia2024continued,
  title={Continued pretraining for domain adaptation of wav2vec2. 0 in automatic speech recognition for elementary math classroom settings},
  author={Attia, Ahmed Adel and Demszky, Dorottya and Ogunremi, Tolulope and Liu, Jing and Espy-Wilson, Carol},
  journal={arXiv preprint arXiv:2405.13018},
  year={2024}
}

@article{lee2025collaborative,
  title={Collaborative learning with artificial intelligence speakers: pre-service elementary science teachers’ responses to the prototype},
  author={Lee, Gyeong-Geon and Mun, Seonyeong and Shin, Myeong-Kyeong and Zhai, Xiaoming},
  journal={Science \& Education},
  volume={34},
  number={2},
  pages={847--875},
  year={2025},
  publisher={Springer}
}

@article{d2024learning,
  title={From learning optimization to learner flourishing: Reimagining AI in Education at the Institute for Student-AI Teaming (iSAT)},
  author={D'Mello, Sidney K and Biddy, Quentin and Breideband, Thomas and Bush, Jeffrey and Chang, Michael and Cortez, Arturo and Flanigan, Jeffrey and Foltz, Peter W and Gorman, Jamie C and Hirshfield, Leanne and others},
  journal={AI Magazine},
  volume={45},
  number={1},
  pages={61--68},
  year={2024},
  publisher={Wiley Online Library}
}

@article{amjad2022advanced,
  title={Advanced learning analytics: aspect based course feedback analysis of MOOC forums to facilitate instructors},
  author={Amjad, Tehmina and Shaheen, Zainab and Daud, Ali},
  journal={IEEE Transactions on Computational Social Systems},
  volume={11},
  number={4},
  pages={4698--4706},
  year={2022},
  publisher={IEEE}
}

@article{takizawa2023using,
  title={Using a topic model to map and analyze a large curriculum},
  author={Takizawa, Peter A},
  journal={Plos one},
  volume={18},
  number={4},
  pages={e0284513},
  year={2023},
  publisher={Public Library of Science San Francisco, CA USA}
}

@inproceedings{rouly2015we,
  title={What are we teaching? Automated evaluation of cs curricula content using topic modeling},
  author={Rouly, Jean Michel and Rangwala, Huzefa and Johri, Aditya},
  booktitle={Proceedings of the Eleventh Annual International Conference on International Computing Education Research},
  pages={189--197},
  year={2015}
}

@article{peyghan2025survey,
  title={A Survey on Non-Intrusive ASR Refinement: From Output-Level Correction to Full-Model Distillation},
  author={Peyghan, Mohammad Reza and Rajabi, Fatemeh and Roudi, Saman Soleimani and Zouashkiani, Saeedreza and Amini, Sajjad and Ghaemmaghami, Shahrokh},
  journal={arXiv preprint arXiv:2508.07285},
  year={2025}
}

@ARTICLE{rao2025,
  author={Ma, Rao and Qian, Mengjie and Gales, Mark and Knill, Kate},
  journal={IEEE Transactions on Audio, Speech and Language Processing}, 
  title={ASR Error Correction Using Large Language Models}, 
  year={2025},
  volume={33},
  number={},
  pages={1389-1401},
  }

@article{pu2023multi,
  title={Multi-stage large language model correction for speech recognition},
  author={Pu, Jie and Nguyen, Thai-Son and St{\"u}ker, Sebastian},
  journal={arXiv preprint arXiv:2310.11532},
  year={2023}
}

@inproceedings{wang-etal-2024-dancer,
    title = "{DANCER}: Entity Description Augmented Named Entity Corrector for Automatic Speech Recognition",
    author = "Wang, Yi-Cheng  and Wang, Hsin-Wei  and Yan, Bi-Cheng  and Lin, Chi-Han  and Chen, Berlin",
    booktitle = "LREC-COLING",
    year = "2024"
}

@inproceedings{huang2024conec,
  title={ConEC: Earnings call dataset with real-world contexts for benchmarking contextual speech recognition},
  author={Huang, Ruizhe and Yarmohammadi, Mahsa and Trmal, Jan and Liu, Jing and Raj, Desh and Garcia, Leibny Paola and Ivanov, Alexei V and Ehlen, Patrick and Yu, Mingzhi and Povey, Dan and others},
  booktitle={LREC-COLING},
  year={2024}
}

@inproceedings{sathyendra2022contextual,
  title={Contextual adapters for personalized speech recognition in neural transducers},
  author={Sathyendra, Kanthashree Mysore and Muniyappa, Thejaswi and Chang, Feng-Ju and Liu, Jing and Su, Jinru and Strimel, Grant P and Mouchtaris, Athanasios and Kunzmann, Siegfried},
  booktitle={ICASSP},
  year={2022}
}

@inproceedings{zhao2019shallow,
  title={Shallow-Fusion End-to-End Contextual Biasing.},
  author={Zhao, Ding and Sainath, Tara N and Rybach, David and Rondon, Pat and Bhatia, Deepti and Li, Bo and Pang, Ruoming},
  booktitle={Interspeech},
  year={2019}
}

@inproceedings{akbik2019flair,
  title={{FLAIR}: An easy-to-use framework for state-of-the-art {NLP}},
  author={Akbik, Alan and Bergmann, Tanja and Blythe, Duncan and Rasul, Kashif and Schweter, Stefan and Vollgraf, Roland},
  booktitle={{NAACL}},
  year={2019}
}

@inproceedings{kwon2023efficient,
  title={Efficient Memory Management for Large Language Model Serving with PagedAttention},
  author={Woosuk Kwon and Zhuohan Li and Siyuan Zhuang and Ying Sheng and Lianmin Zheng and Cody Hao Yu and Joseph E. Gonzalez and Hao Zhang and Ion Stoica},
  booktitle={Symposium on Operating Systems Principles},
  year={2023}
}

@inproceedings{he2019streaming,
  title={Streaming end-to-end speech recognition for mobile devices},
  author={He, Yanzhang and Sainath, Tara N and Prabhavalkar, Rohit and McGraw, Ian and Alvarez, Raziel and Zhao, Ding and Rybach, David and Kannan, Anjuli and Wu, Yonghui and Pang, Ruoming and others},
  booktitle={ICASSP},
  year={2019}
}

@inproceedings{chang2021context,
  title={Context-aware transformer transducer for speech recognition},
  author={Chang, Feng-Ju and Liu, Jing and Radfar, Martin and Mouchtaris, Athanasios and Omologo, Maurizio and Rastrow, Ariya and Kunzmann, Siegfried},
  booktitle={ASRU},
  year={2021},
}

@article{mohri2002weighted,
  title={Weighted finite-state transducers in speech recognition},
  author={Mohri, Mehryar and Pereira, Fernando and Riley, Michael},
  journal={Computer Speech \& Language},
  year={2002}
}

@article{gong2024advanced,
  title={Advanced long-content speech recognition with factorized neural transducer},
  author={Gong, Xun and Wu, Yu and Li, Jinyu and Liu, Shujie and Zhao, Rui and Chen, Xie and Qian, Yanmin},
  journal={IEEE/ACM Transactions on Audio, Speech, and Language Processing},
  year={2024},
  publisher={IEEE}
}

@inproceedings{radford2023robust,
  title={Robust speech recognition via large-scale weak supervision},
  author={Radford, Alec and Kim, Jong Wook and Xu, Tao and Brockman, Greg and McLeavey, Christine and Sutskever, Ilya},
  booktitle={ICML},
  year={2023}
}

@inproceedings{kannan2018analysis,
  title={An analysis of incorporating an external language model into a sequence-to-sequence model},
  author={Kannan, Anjuli and Wu, Yonghui and Nguyen, Patrick and Sainath, Tara N and Chen, Zhijeng and Prabhavalkar, Rohit},
  booktitle={ICASSP},
  year={2018}
}

@inproceedings{pundak2018deep,
  title={Deep context: end-to-end contextual speech recognition},
  author={Pundak, Golan and Sainath, Tara N and Prabhavalkar, Rohit and Kannan, Anjuli and Zhao, Ding},
  booktitle={Spoken language technology workshop (SLT)},
  year={2018},
}

@article{kuchaiev2019nemo,
  title={Nemo: a toolkit for building ai applications using neural modules},
  author={Kuchaiev, Oleksii and Li, Jason and Nguyen, Huyen and Hrinchuk, Oleksii and Leary, Ryan and Ginsburg, Boris and Kriman, Samuel and Beliaev, Stanislav and Lavrukhin, Vitaly and Cook, Jack and others},
  journal={arXiv preprint arXiv:1909.09577},
  year={2019}
}

@article{philips2000doublemetaphone,
  author = {Philips, Lawrence},
  title = {The Double Metaphone Search Algorithm},
  journal = {C/C++ Users Journal},
  year = {2000}
}

@inproceedings{ott2019fairseq,
  title = {fairseq: A Fast, Extensible Toolkit for Sequence Modeling},
  author = {Myle Ott and Sergey Edunov and Alexei Baevski and Angela Fan and Sam Gross and Nathan Ng and David Grangier and Michael Auli},
  booktitle = {NAACL-HLT Demonstrations},
  year = {2019}
}

@inproceedings{blair2024jrc,
  title={JRC-Names-Retrieval: A Standardized Benchmark for Name Search},
  author={Blair, Philip and Bar, Kfir},
  booktitle={LREC-COLING},
  year={2024}
}

@article{brown2020language,
  title={Language models are few-shot learners},
  author={Brown, Tom and Mann, Benjamin and Ryder, Nick and Subbiah, Melanie and Kaplan, Jared D and Dhariwal, Prafulla and Neelakantan, Arvind and Shyam, Pranav and Sastry, Girish and Askell, Amanda and others},
  journal={Neural information processing systems},
  year={2020}
}

\balancecolumns
\end{document}